# End-to-End Fingerprints Liveness Detection using Convolutional Networks with Gram module

Eunsoo Park[1*], Xuenan Cui[1], Weonjin Kim[1], Hakil Kim[1]
[1] Information and Communication Engineering, Inha University, 100 inharo, Nam-gu Incheon 22212, South Korea
es.park@modulabs.co.kr (Eunsoo Park)

**Abstract:** This paper proposes an end-to-end CNN(Convolutional Neural Networks) model that uses gram modules with parameters that are approximately 1.2MB in size to detect fake fingerprints. The proposed method assumes that texture is the most appropriate characteristic in fake fingerprint detection, and implements the gram module to extract textures from the CNN. The proposed CNN structure uses the fire module as the base model and uses the gram module for texture extraction. Tensors that passed the fire module will be joined with gram modules to create a gram matrix with the same spatial size. After 3 gram matrices extracted from different layers are combined with the channel axis, it becomes the basis for categorizing fake fingerprints. The experiment results had an average detection error of 2.61% from the LivDet 2011, 2013, 2015 data, proving that an end-to-end CNN structure with few parameters that is able to be used in fake fingerprint detection can be designed.

## 1. Introduction

With the rise of authentication systems using fingerprint recognition in smartphones, studies on fingerprint recognition technology for mobile devices are actively underway. Authentication systems that use fingerprint recognition is presently evaluated as an authentication method with outstanding growth thanks to ease of use and economic advantages of low setup costs. However, security issues are also garnering much interest following the popularity of authentication systems that use fingerprints. For example, a doctor in Brazil who helped an absent colleague clock in by using a fake fingerprint was imprisoned in March 2013. When the iPhone 5S was released in September 2013 equipped with smartphone unlocking and payment functions using fingerprints, it took only a few days after its release to prove that authentication can easily be achieved using fake fingerprints made of adhesives [1]. Fingerprints are easily forged using silicon, gelatin, clay, etc. Therefore, if safety or security is an important factor, fingerprint recognition systems will require the ability to assess whether or not a fingerprint is fake.

In terms of software-based methods, there are a growing number of studies using the Convolutional Neural Networks (CNN) with high categorization performance for detecting fake fingerprints [2]. Because the majority of CNN studies focus on identifying and categorizing objects from common photos that can be found on the web, there are inevitable differences from fingerprint images and good performance will not be achieved simply applying CNN. The following are issues that occurred upon applying CNN for fake fingerprint detection in the past.

- The structure that was used to categorize simple images was utilized as is without considering the characteristics of fake fingerprints.
    - Most trained CNN model parameters were obtained from pictures of nature. The parameters acquired here can not be regarded as optimized parameters for fake fingerprint detection.
- Most of the focus is placed on increasing classification performance rather than considering the size of parameters in the CNN model.
    - Many studies have shown that better performance is achieved when a deeper and wider network is utilized [3, 4]. However, when considering the expandability of the fake fingerprint detection method, the model must include only a small number of parameters so that it can run in smartphones with small memory sizes.
- Due to various preprocessing steps such as cropping through segmentation or reducing images to match the input size configured upon applying CNN, a complete end-to-end type fake fingerprint detection network cannot be proposed.

In order to resolve the above issues, this paper proposes an end-to-end CNN model using gram matrices with parameters that are approximately 1.2MB that can operate regardless of the fingerprint input size. The total number of parameters in the proposed CNN model is 308,554, and if one parameter is 4bytes, it will be 1.2MB in size. Gram modules were implemented in order to detect the appropriate characteristics for fake fingerprint detection and to configure a network that is irrelevant to the size of fingerprint inputs. The proposed CNN model takes the fingerprint image input as is without preprocessing, then differentiates between fake and live fingerprints. The experiment results had an average detection error of 2.61% for fake fingerprints, showing the feasibility of building an end-to-end CNN model with high performance despite having only few parameters.

This paper is structured as follows. Chapter 2 introduces other studies relevant to fake fingerprint detection and Chapter 3 explains the proposed method. Chapter 4 describes the experimental method and the results and Chapter 5 discusses the conclusion and direction for future studies.

## 2. Related Studies

According to the categorization proposed by Coli[5], fake fingerprint detection can be divided into hardware based methods and software based methods. Hardware based methods use additional hardware for extracting physical characteristics from the human body. These methods can make more accurate detections compared to software methods, but they are more expensive due to the extra sensor. Software based methods detect fake fingerprints by using an algorithm. They are cheaper than hardware methods since there is no need for additional hardware. Most of the algorithms utilize physical data such as the size of fingerprint ridges, density, continuity, etc. Studies that applied CNN have recently been conducted.

Because the method proposed in this paper is a software based fake fingerprint detection method, key studied related to fake fingerprint detection using texture as the fake fingerprint characteristics, similar to this study, will be examined below. Nikam and Agarwa[6] proposed a method that combined the local binary pattern (LBP) and wavelet transformation. The LBP histogram is used when analyzing textures and wavelets are used when analyzing the frequency characteristics and direction information of ridges. Nikan and Agarwa furthered their study by applying wavelets and GrayLevel Co-occurrence Matrix (GLCM) to propose accessibility that extracts texture characteristics[7]. This method uses the Principal Component Analysis (PCA) and Sequential Forward Feature Selection (SFFS) methods to reduce the dimensionality of certain groups.

Coli et al.[8] proposed a method that categorizes fake fingerprints after applying Fourier transformations to fingerprint images by using the minor characteristics of live fingerprints that are not easily visible in fake fingerprints due to the rough surface and inconsistency of ridges. High frequency energy (HFE) were defined in order to measure only certain frequencies, and these were used in analysis.

Marasco[9] proposed a fake fingerprint detection method that used texture as the characteristic. Texture characteristics include characteristics that occur through signal processing such as the size of sweat glands and fingerprint static, statistical characteristics such as dispersion and information quantity, and image gray level characteristics. Galbally et al.[10] used a similar approach, but extracted texture characteristics using the Gabor filter and used these for fake fingerprint detection. Gottschlich et al.[11] improved the Histogram of Oriented Gradients (HOG) and Scale Invariant Feature Transform (SIFT), and used these to propose and apply a Histograms of Invariant Gradients (HIG) that was modified to fit fingerprint ridge textures.

Ghiani et al.[12] supplemented the LBP method and applied a Local Phase Quantization (LPQ) that is strong against rotation. LPQ is typically used for low frequency component analyses due to observations that low frequency component analyses include characteristics that are advantageous for differentiating between fake and live fingerprints. Gragnaniello et al.[13] used this method with the Weber Local Descriptor(WLD) and LPQ method to show that better performance can be achieved. Jia[14] applied size modifications of filters used in LBP and various linear filters in fingerprint images, then applied the results to LBP again to extract characteristics, and applied these to fake fingerprint images to prove that better fake fingerprint detection can be achieved compared to other existing studies.

New fake fingerprint detection methods using CNN can be divided into methods that use fingerprints by cutting them up into patches[15, 16, 17] and methods that apply CNN after modifying the size of images to fit 224×224 or 227×227 as in general classification methods[18, 19, 20]. Studies that use fingerprint patches divided fingerprints into small 16×16 or 32×32 patches before applying CNN, then processed the results to determine whether or not a fingerprint is fake. In studies that applied methods that are used for general classifications as is for fingerprints, fingerprints were cut up to fit the network's input size, then expanded or minimized before applying CNN. Nogueira et al.[18, 19] used VGG-19[21] and Marasco et al.[20] used GoogLeNet[3] to classify fake fingerprints. Fig. 1 (a) and (b) compare the method that uses patches and the method that cuts up fingerprints for input into CNN.

(a) and (b) in Fig. 1 both require preprocessing before applying CNN. The proposed method proceeds the same way as (c) in Fig. 1, and it uses the fewest parameters out of all CNN models that have been used for fake fingerprint detection to the present.

The proposed method does not simply use pre-trained models as in existing methods, but instead uses gram matrices in order to extract fingerprint texture information. Gram matrices are used in style transfers proposed by Gatys et al.[22, 23] and are applied when extracting characteristics from an aesthetic image.

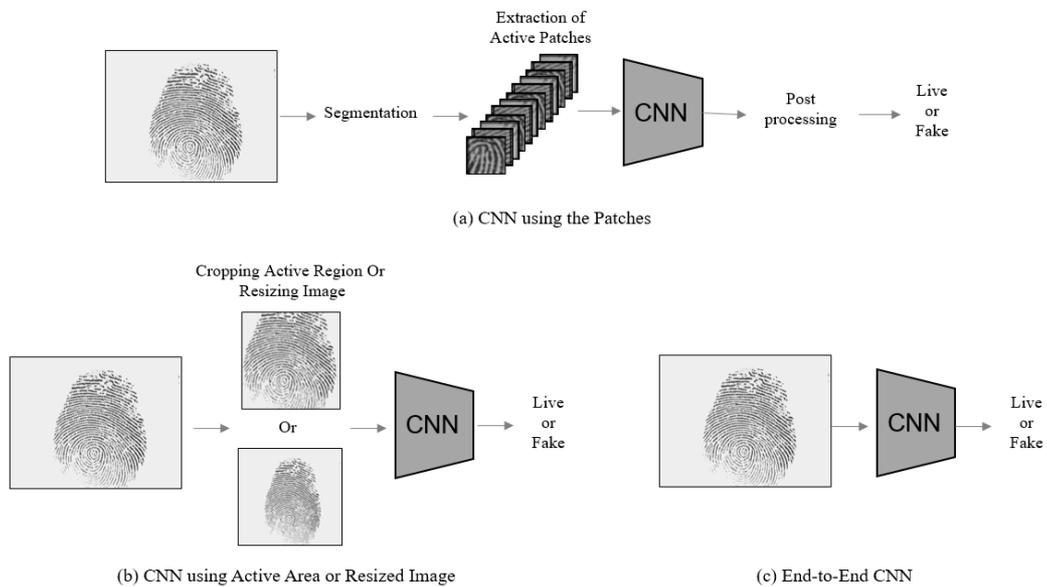

**Fig. 1.** Methodology of applying CNN in fake fingerprint detection. (a) Using fingerprint patches; (b) Changing fingerprints for CNN input; (c) Using original fingerprint as is

Gatys proposed a method that changes the style of a picture into the style that was used in artists' pictures while retaining the content of the input image. Gram matrices are implemented in order to extract styles from artists' pictures, and this involves calculating the correlation coefficient for each CNN tensor channel and turning it into matrix form. The difference between the gram matrix acquired from artists' pictures and the gram matrix of the image to be changed is turned into a loss function, and the pixels of the input image are updated instead of updating parameters. When the input image is repeatedly updated as explained above while retaining the content from the original input image, the input image can be transformed into a picture with the unique style of artists. The proposed method assumed the artists' styles are texture characteristics and used gram matrices that were used to extract these in fake fingerprint detection.

## 3. Proposed Method

Models that learned classifiers after extracting texture data by using LBP or Gabor filters often demonstrated favorable performance in fake fingerprint research[6, 7, 10, 12]. This section proposes a method that uses gram matrices to extract and apply fake fingerprint texture information from CNN. The proposed method extracts gram matrices from fingerprints based on the style transfer idea from Gatys[22], then uses these to detect fake fingerprints. In other words, the style defined by the style transfer method is assumed as the image's texture, and gram matrices are applied in order to extract these textures.

### 3.1. Gram Matrix

If a gram matrix has a 3-dimensional tensor with a height, width, and channel size of $H \times W \times C$ as shown in Fig. 2, this is turned into a 2-dimensional $(H * W) \times C$ matrix before creating a correlation matrix for each row component. To generalize the process of creating gram matrices, the 2-dimensional matrix transformation results of $(H*W) \times C$, which is the middle stage in Fig. 2, is assumed to be $F^l$. Here, $l$ is the $l$th layer of the neural network that was inserted for the purpose of generalization. Assuming that $F_i^l$ is the $i$th row vector of Matrix $F^l$, each factor $G_{i,j}^l$ of the 2-dimensional gram matrix of layer $l$ is calculated through the inner product as shown in Formula (1).

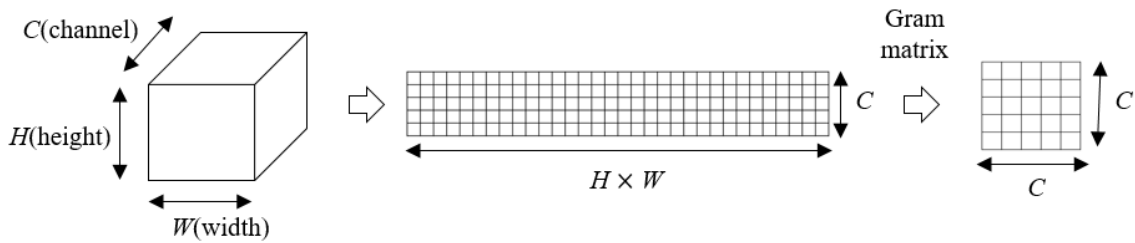

**Fig. 2.** Process of generating a gram matrix.

$$G_{i,j}^l = F_i^l \cdot F_j^l \qquad (1)$$

Gram matrix $G^l$ is a symmetric matrix and was used to extract the style of artistic pictures in the study conducted by Gatys[22]. Style characteristics may be interpreted as overall textures that are not dependent on a location through gram calculation, and this characteristic can also be used in fake fingerprint detection.

### 3.2. CNN Structure Using Gram Modules

Gram modules are defined as shown in Fig. 3 in order to include them in CNN. Gram modules are composed of 1×1 convolution layers, tanh nonlinear activation function, and gram layers. Gram modules adjusts the number of 1×1 convolution filters to create gram matrices of the desired size. For the sake of convenience, the gram module with $K$ 1×1 filters is defined as Gram-$K$. Fig. 3 shows an example of the $K{\times}K$ gram matrix that is obtained through the results of Gram-$K$. A CNN structure that tolerates any size input is achieved through this.

The proposed method used a fire module [24] that is used in SqueezeNet to design a CNN with only a few parameters. The fire module is composed of squeeze layers and expand layers as shown in Fig. 4. In the squeeze layer, the number of 1×1 filters are minimized to reduce the number of input channels, then tensors are outputted and used as the input for expand layers. Since the number of channels has decreased, the number of parameters for filters used can also be reduced. Expand layers are composed of 1×1 and 3×3 filters. In the proposed network, the number of 1×1 filters and 3×3 filters used in the expand layer are identical. The number of 1×1 filters used in the squeeze layer can be calculated by multiplying 0.125 to the number of filters in the expand layer. Fig. 5 shows the proposed network that uses fire modules and gram modules.

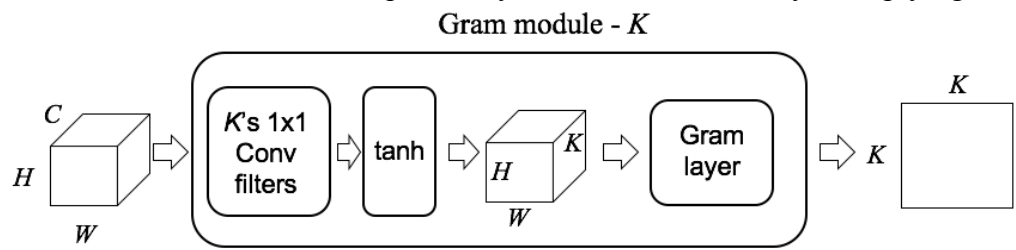

**Fig 3.** The Gram-K module is used to acquire the $K{\times}K$ gram matrix.

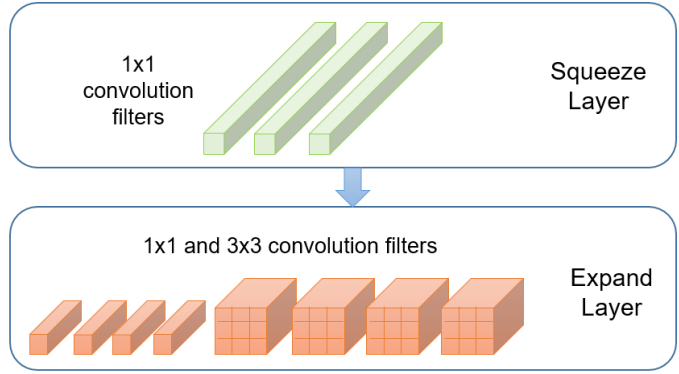

**Fig 4.** Fire Module Structure.

96 7×7 filters with stride 2 was used for the first layer in Fig. 5, conv1-96 / 2. The last convolution layer, conv7-2, is 2 1×1 filters. maxpool / 2 in Fig. 5 refers to max pooling with stride 2 at a size of 3×3, and avgpooling refers to global average pooling. fire2-128 in Fig. 5 means there are 128 expand layer filters. Hence, the number of 1×1 and 3×3 filters in the expand layer becomes 64 and the number of 1×1 filters in the squeeze layer is 16. Gram-128 is the gram module, and the size of the outputted tensor is 128×128×1 . The 3 gram matrices that are acquired from each Gram-128 module become stacked in the channel direction take a 128×128×3 shape, which becomes the input for the fire module once again. Leaky ReLU[25] shown in Formula (2) was used as the activation function that is used in the network, and the value of $a$ was set to 0.3. Before the activation function was applied to all layers, a batch normalization layer[26] was added.

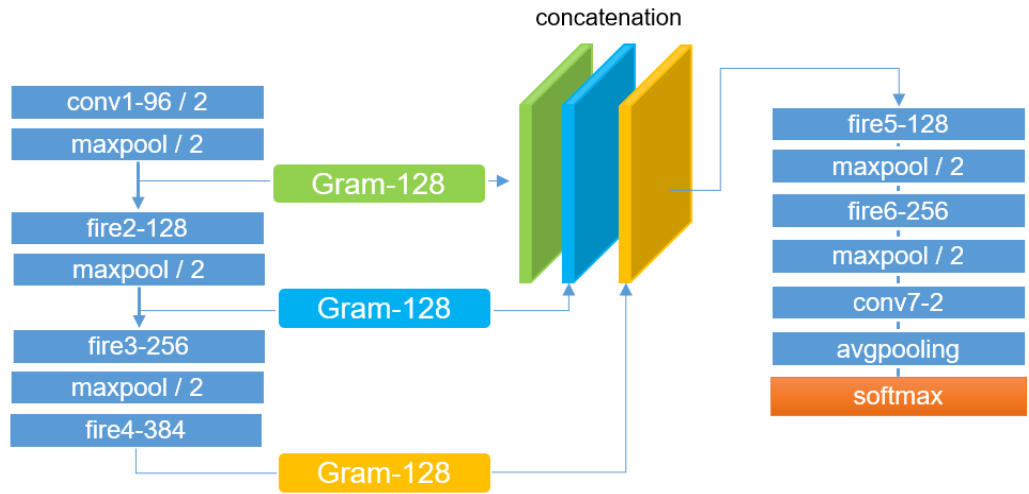

**Fig 5.** Proposed CNN structure that uses gram modules and fire modules.

By applying gram modules, the proposed model can process inputs of any size without requiring segmentation or other processes, and can operate through an end-to-end method. The total number of parameters in the network was 308,554, and size was approximately 1.2MB. Table 1 shows the detailed structure of the CNN model using gram matrices when the size of the fingerprint image input is $K \times K$. The 128×128×3 tensor that was combined by passing the gram modules has global texture characteristics that appear across the fingerprint. Despite them being global characteristics, the proposed method placed 2 fire modules by inputting a 128×128×3 tensor to share parameters and reduce the number of parameters.

$$f(x) = \max(x, ax) \qquad (2)$$

**Table 1**
Structure of the CNN model based on gram matrices.

| Laer name / type | Output size | Fiter size / stride (if not a fire layer) | depth | $s_{1\times1}$ (#1×1 squeeze) | $e_{1\times1}$ (#1×1 expand) | $e_{3\times3}$ (#3×3 squeeze) | # of paramters |
|---|---|---|---|---|---|---|---|
| input image | $K \times K \times 1$ | | | | | | |
| conv1 | $\frac{K}{2} \times \frac{K}{2} \times 96$ | 7×7/2 (×96) | 1 | | | | 4,800 |
| maxpool1 | $\frac{K}{4} \times \frac{K}{4} \times 96$ | 3×3/2 | 0 | | | | |
| gram1 | 128×128×1 | 1×1 (×128) | 1 | | | | 12,416 |
| fire2 | $\frac{K}{4} \times \frac{K}{4} \times 128$ | | 2 | 16 | 64 | 64 | 11,920 |
| maxpool2 | $\frac{K}{8} \times \frac{K}{8} \times 128$ | 3×3/2 | 0 | | | | |
| gram2 | 128×128×1 | 1×1 (×128) | 1 | | | | 16,512 |
| fire3 | $\frac{K}{8} \times \frac{K}{8} \times 256$ | | 2 | 32 | 128 | 128 | 45,344 |
| maxpool3 | $\frac{K}{16} \times \frac{K}{16} \times 256$ | 3×3/2 | 0 | | | | |
| fire4 | $\frac{K}{8} \times \frac{K}{8} \times 384$ | | 2 | 48 | 192 | 192 | 104,880 |
| gram3 | 128×128×1 | 1×1 (×128) | 1 | | | | 49,280 |
| concatenation | 128×128×3 | | 0 | | | | |
| fire5 | 128×128×128 | | 2 | 16 | 64 | 64 | 10,432 |
| maxpool5 | 63×63×128 | 3×3/2 | 0 | | | | |
| fire6 | 31×31×256 | | 2 | 32 | 128 | 128 | 45,344 |
| maxpool6 | 31×31×256 | 3×3/2 | 0 | | | | |
| conv10 | 31×31×2 | 1×1/1(×2) | 1 | | | | |
| avgpool10 | 1×1×2 | 31×31/1 | 0 | | | | |
| Total #of parameters including batch normalization layers : 308,554 (about 1.2MB) | | | | | Total # of parameters | | 301,442 |

## 4. Experiment Results

### 4.1. Experimental Data

The data used to evaluated the performance of the proposed method included LivDet2011[27], LivDet2013[28], and LivDet2015[29]. Only Italdata sensor data and Biometrika sensor data acquired via the non-cooperative method were used for LivDet2013. Since the swipe sensor for LivDet2013 data acquires fingerprints by swiping from top the bottom, the images that were acquired were vastly different from existing data, and because the Crossmatch sensor for LivDet2013 showed issues upon acquiring fingerprints, these were eliminated from the data used for analysis. Table 2 shows the LivDet data used for the experiment.

**Table 2**
LivDet2011 data used in the experiment.

| DBs | Sensor | Size | DPI | # of training(Live/Fake)/ # of testing (Live/Fake) | # of Fake materials | Example Image (real image ratio) Live | Fake |
|---|---|---|---|---|---|---|---|
| LivDet2011 | Biometrika | 312×372 | 500 | (1000/1000)/(1000/1000) | 5 | 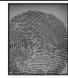 | 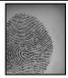 |
|  | Digital Persona | 355×391 | 500 | (1000/1000)/(1000/1000) | 5 | 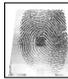 | 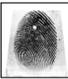 |
|  | Italdata | 640×480 | 500 | (1000/1000)/(1000/1000) | 5 | 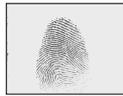 | 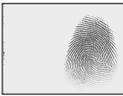 |
|  | Sagem | 352×384 | 500 | (1000/1000)/(1000/1000) | 5 | 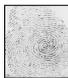 | 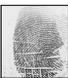 |
| LivDet2013 | Biometrika | 312×372 | 569 | (1000/1000)/(1000/1000) | 5 | 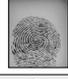 | 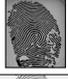 |
|  | Italdata | 640×480 | 500 | (1000/1000)/(1000/1000) | 5 | 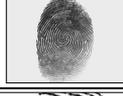 | 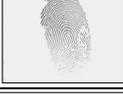 |
| LivDet2015 | Biometrika | 1000×1000 | 1000 | (1000/1000)/(1000/1000) | 4 | 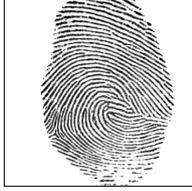 | 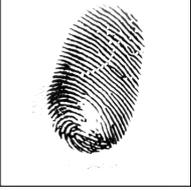 |
|  | Crossmatch | 800×750 | 500 | (1510/1473)/(1500/851) | 3 | 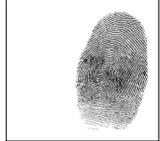 | 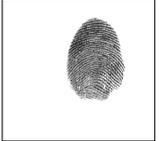 |
|  | Digital Persona | 252×324 | 500 | (1000/1000)/(1000/1000) | 4 | 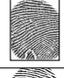 | 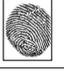 |
|  | Green Bit | 500×500 | 500 | (1000/1000)/(1000/1000) | 4 | 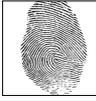 | 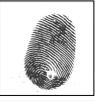 |

Because LivDet2015 data included data composed of materials that were not used for training, only the fake fingerprint material used for training were included in the test data set. Hence, there were differences in the number of live fingerprints and fake fingerprints from the LivDet2015 Crossmatch sensor. The fingerprint images shown as examples in Table 2 are the same size ratio as the actual fingerprints. Italdata sensor and Crossmatch sensors that typically include much of the background require a preprocessing such as segmentation. Moreover, since the size of fingerprints from the LivDet2015 Biometrika sensor is 1000×1000, there was concern that too much data would be lost upon reducing the size down to 256×256 for CNN input. Even if they were used after certain areas were cut off, only an extremely small part of the fingerprint would be inputted and used. Such concerns are irrelevant to the proposed method as it runs regardless of the size of the input image, thus all the data in Table 2 could be processed using the end-to-end method through the same network structure.

### *4.2.* Experimental Environment and Experiment Results

The GPU used in the experiment was NVIDA GTX 1080. In terms of the parameter used for training, the learning rate was 0.0005, the batch size was 8, and the number of epoch was 80. Adamax[30] was used for the optimizer and 10% of the LivDet training data was used for validation set. Fingerprints from other people were used for validation data whenever possible and data distribution was made identical to the training data. Random horizontal flip was applying per batch of training data, and the loss value of verification for each epoch was confirmed. If loss value did not decrease for 4 epochs, the learning rate was half. Table 3 shows the hyperparameters used in the experiment. Table 4 shows the ACE(Average Classification Error) of fake fingerprint detection for evaluating the performance of fake fingerprint detection from the proposed model that uses gram modules.

**Table 3**

Experimental environment and hyperparameters for learning.

| | |
|---|---|
| GPU | NVIDIA GTX 1080 |
| Learing rate | 0.0005 (making it half when validation loss does not decrease for 4 epoch) |
| Batch Size | 8 |
| Epoch | 80 |
| Optimizer | Adamax |
| Validation data | 10% of training data |

**Table 4**

Comparison of performance between the method using gram modules and state-of-the-art method.

| DB | Sensor | State Of The Art | Gram-128 model (ACE) | Gram-128 model (Augmentation) (ACE) | Processing Time (ms) |
|---|---|---|---|---|---|
| 11 | Biometrika | 3.5 [16] | 2.75 | 4.95 | 13 |
|  | Digital Persona | 0 [16] | 0.55 | 2 | 13 |
|  | Italdata | 0 [16] | 5 | 4.8 | 21 |
|  | Sagem | 0 [16] | 1.5 | 2.56 | 13 |
| 13 | Biometrika | 0.8 [19] | 0.85 | 0.7 | 13 |
|  | Italdata | 0 [16] | 1.25 | 0.9 | 21 |
| 15 | Biometrika | 5.6 [29] | 4.1 | 3.75 | 52 |
|  | Crossmatch | 1.53 [29] | 0.27 | 3.4 | 35 |
|  | Digital Persona | 6.35 [29] | 8.5 | 7 | 12 |
|  | Green Bit | 3.9 [29] | 1.35 | 2.46 | 17 |
|  | Average | **2.34** | **2.61** | **3.25** | **21** |

*ACE* is calculated as shown in Formula (3). *Ferrlive* in Formula (3) refers to the ratio of live fingerprints that were incorrectly classified as fake fingerprints while *Ferrfake* refers to the ratio of fake fingerprints that were incorrectly categorized as live fingerprints.

$$ACE = (Ferrlive + Ferrfake)/2 \qquad (3)$$

Table 4 shows that the average detection error rate for the proposed method was 2.61%, which is not so different from the state-of-the-art's 2.34%. The state-of-the-art results in Table 4 are the best results obtained from the patch based CNN method from Wang[16] and the VGG model from Nogueira[19] to the present. The *ACE* of Wang's method is very low but they did not mention about validation data set. Since there are not many results for LivDet2015's state-of-the-art performance in literature thus far, the results of using Nogueira's CNN, which was the winner of LivDet 2015 competition, were used instead.

Data augmentation was applied to the results in Table 4 and these are the results of applying random vertical flip in addition to horizontal flip. The effect of data augmentation differs according to sensor, but performance is typically worse than when data augmentation is not applied. Processing time, which is an obstacle when differentiating fake fingerprints, is proportional to the size of the input image. The proposed method shows an average of 21ms, demonstrating that fake fingerprints can be differentiated at a considerably high speed. Fig. 6 is a graph of the accumulated results of ACE from Table 4.

The DET(Detection Error Tradeoff) curve of the Gram-128 model in Table 4 is shown in Fig. 7. Fig. 7(a) shows that *Ferrlive* hardly decreases for the Italdata sensor. The value of *ACE* for the 2011 Italdata sensor in Table 4 was actually calculating by setting $Ferrlive$ to 10% and the value of $Ferrfake$ to 0%. Fig. 7(b) is an expanded view of error range to [0%.20%] so that the changing trend can be seen more clearly.

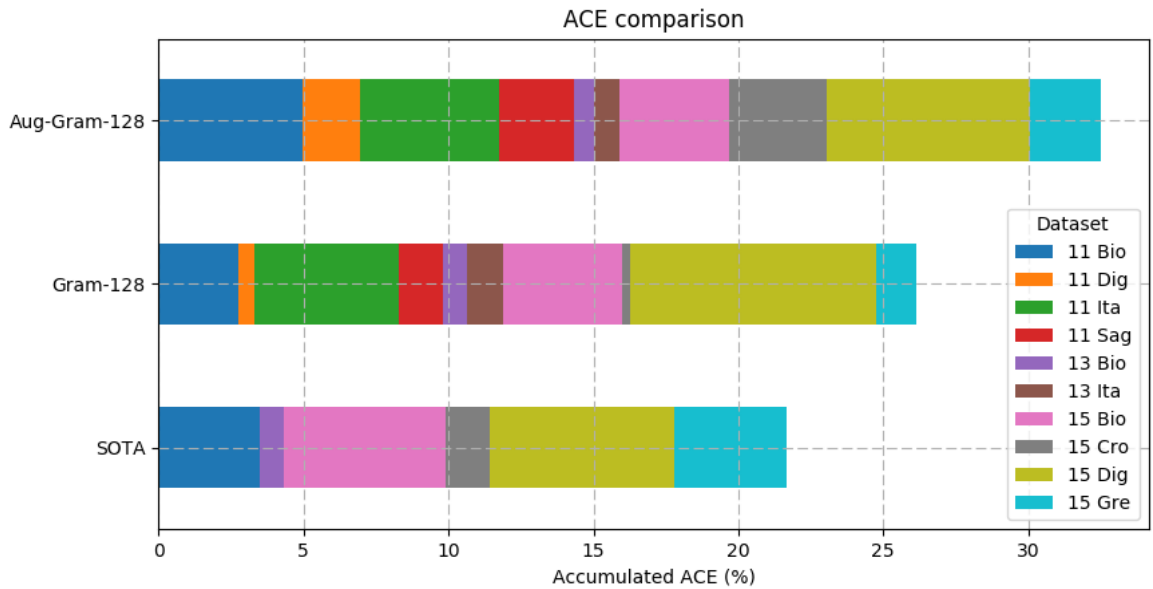

**Fig 6.** Comparison of accumulated ACE for the proposed method.

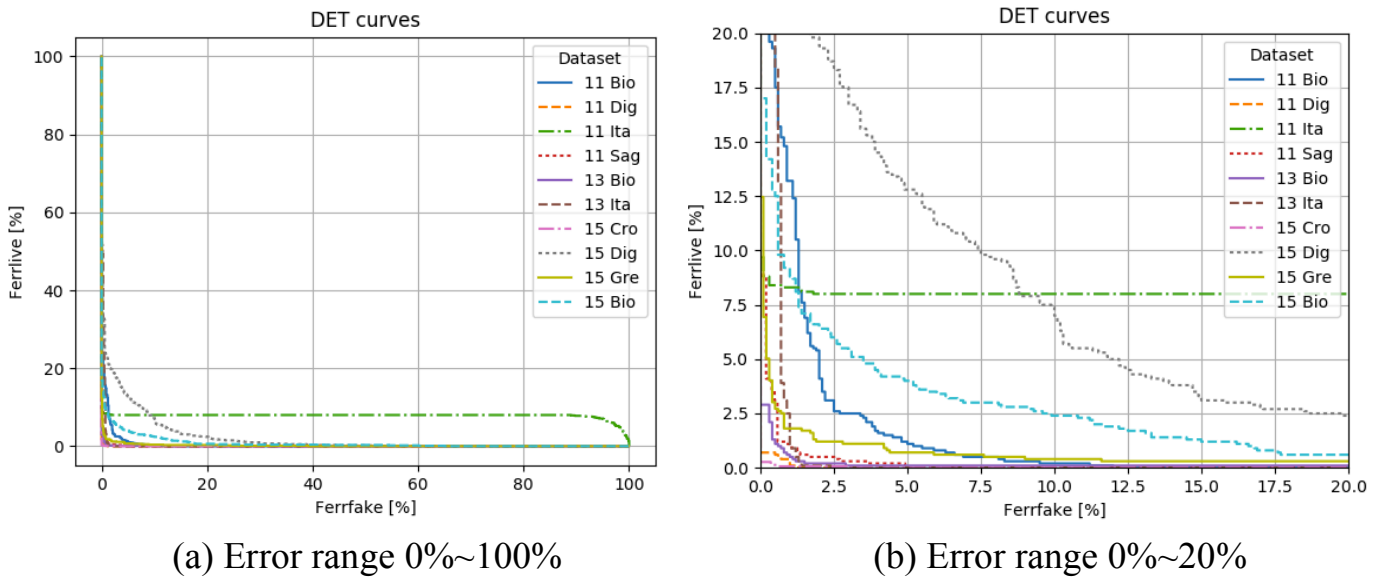

(a) Error range 0%~100%  (b) Error range 0%~20%

**Fig 7.** DET curve of the Gram-128 model. (a) All error shown from 0-100%, (b) Error shown from 0%-20.

Fake fingerprints that were made of materials that were not used during training are included in the test data for LivDet 2015. The unknown fake fingerprints materials that were not present during training included liquid ecoflex and RTV for the Biometrika, Digital Persona, and Green Bit sensors, and OOMOO and gelatin

**Table 5**
Generalization performance evaluation for the proposed method (detection rate: %)

| rank | Biometrika | | CrossMatch | | Digital Person | | GreenBit | |
|---|---|---|---|---|---|---|---|---|
| 1 | unina | 98.6 | **Gram** | **100** | unina | 99.4 | **Gram** | **96.2** |
| 2 | titanz | 95 | COPILHA | 98.32 | nogueira | 94 | unina | 96 |
| 3 | nogueira | 94.2 | anonym | 95.98 | **Gram** | **90.6** | nogueira | 92.6 |
| 4 | hbirkholz | 93.8 | nogueira | 95.98 | UFPE I | 85.4 | jinglian | 92.2 |
| 5 | jinglian | 93.2 | jinglian | 88.44 | hbirkholz | 85.2 | titanz | 87.6 |
| 6 | CSI_MM | 88.6 | unina | 86.1 | titanz | 84.4 | hectorn | 87.2 |
| 7 | anonym | 85.4 | hbirkholz | 81.41 | jinglian | 80.6 | anonym | 86.4 |
| 8 | hectorn | 83.2 | titanz | 80.74 | CSI | 75.8 | UFPE II | 83.6 |
| 9 | **Gram** | **82** | hectorn | 76.55 | CSI_MM | 73.2 | CSI_MM | 82.2 |
| 10 | CSI | 80.8 | CSI_MM | 70.18 | UFPE II | 72.4 | hbirkholz | 81.6 |
| 11 | UFPE II | 72 | CSI | 69.68 | hectorn | 70.8 | CSI | 76 |
| 12 | UFPE I | 58.6 | UFPE I | 52.43 | anonym | 70.8 | COPILHA | 75.6 |
| 13 | COPILHA | 57.2 | UFPE II | 45.9 | COPILHA | 69.4 | UFPE I | 63 |

for the Crossmatch sensor. Table 5 shows the results of the accuracy of fake fingerprint detection using only this data. Table 5 also shows the results of participants regarding unknown materials from the LivDet2015

Competition alongside the results of the proposed method for a comparison. Based on the experiment results, extremely good performance was observed for the proposed method even regarding unknown material data excluding the Biometrika sensor.

## 5. Conclusion and future works

This paper proposed an end-to-end CNN model using gram modules with parameters that are approximately 1.2MB in size that can operate regardless of the size of the fingerprint input. The proposed method assumes that texture is the most appropriate characteristic in fake fingerprint detection, and proposed gram modules for extracting textures from the CNN network. Instead of simply building a deep and wide network for high performance fake fingerprint detection, fire modules were used as the basic structure of CNN in order to build a network with just a few parameters. Tensors that passed the gram modules in the proposed network were recombined and applied for fake fingerprints detection, and the experiment results showed that detection performance was on par with state-of-the-art results. There were a total of 308,554 parameters in the proposed network (approximately 1.2MB), which is an extremely small parameter count, while still being able to fully operate end-to-end.

The influence of LivDet data is large based on how CNN can now be utilized in fake fingerprint detection research. Building a fake fingerprint database is much more difficult than acquiring data using general object categories. Even if a database with many fake fingerprints is built, a fake fingerprint attack using new material can occur at any time. In the future, fake fingerprints will be learned through techniques such as the Generative Adversarial Network for better generalization performance, and a method that learns LivDet data with less sensor dependency instead of training for each sensor will be studied.